\documentclass[11pt]{article}

\usepackage[margin=1in]{geometry}
\usepackage{amsmath,amssymb,amsthm}
\usepackage{booktabs}
\usepackage{multirow}
\usepackage{graphicx}
\usepackage{array}
\usepackage{float}
\usepackage{xcolor}
\usepackage[protrusion=true,expansion=false]{microtype}
\usepackage[numbers,sort&compress]{natbib}
\usepackage[colorlinks=true,linkcolor=blue,citecolor=blue,urlcolor=blue]{hyperref}
\newcommand{\humaneval}{\textsc{HumanEval}}

\newcommand{\oci}{\textsc{OpenCodeInstruct}}
\newcommand{\curated}{\textsc{Curated}}
\newcommand{\lit}{\textsc{Literature}}
\newcommand{\base}{\textsc{Base}}
\newcommand{\ado}{\textsc{Ado}}
\newcommand{\passk}[1]{\textrm{pass@}#1}

\newtheorem{definition}{Definition}
\newtheorem{remark}{Remark}

\title{Conditional Transfer from Controlled Pretraining Mixtures to Code}

\author{Ohad Rubin}

\date{}

\begin{document}
\maketitle

\begin{abstract}
Synthetic tasks are increasingly used both as probes of language-model
capability and as pretraining data. These two uses are usually justified by
the same evidence: a task whose loss falls during training is treated as
informative, and a task whose loss falls faster when it is sampled more is
treated as worth sampling. We separate three signals that this reasoning
conflates. A task is \emph{diagnostic} when its loss tracks global pretraining
progress; it is \emph{teachable} when its loss responds to its own token budget;
and a data source \emph{transfers} when including it improves a downstream
target. We study these signals in a controlled pretraining setting in which
$70\%$ of the corpus is fixed general Python and the remaining $30\%$ is a
simplex over three source families: \oci{}, a curated suite of $12$
code-adjacent synthetic tasks, and $15$ literature-derived probe tasks. Across
a task-budget sweep we detect teachability for $14$ of $27$ tasks, with a sharp
asymmetry between the two synthetic families ($10/12$ curated versus $4/15$
literature-derived).

Teachability and downstream transfer give different rankings. On the
mixture-simplex edge between the curated suite and \oci{}, \humaneval{} $\passk{20}$ after a
fixed fine-tuning stage rises from $15.9$ at pure curated data to its highest
observed value, $22.6$, at a mixture that is $75\%$ \oci{}, then falls to $19.5$
at pure \oci{}. Curated synthetic data therefore has \emph{conditional} value:
it contributes as a limited share of a mixture that a target-aligned source
still dominates. Finally, a loss-based adaptive scheduler exposes the
mismatch between residual loss reducibility and downstream transfer. Across
three $60$k-step free-ratio runs, \ado{} drives the \oci{} share below $5\%$
within the first $5$k steps and to $1.2$--$1.4\%$ by the end of training, and
underperforms its matched fixed-mixture controls by $2.4$--$11.0$ percentage
points. Optimizing near-term task-loss reduction moves the mixture away from the
region that transfers.
\end{abstract}

\section{Introduction}

A pretraining corpus is a mixture, and choosing the mixture is now a first-class
design problem
\citep{xie2023doremi,fan2024doge,chen2023skillit,jiang2024ado}.
Two questions get asked of a candidate source. The first is diagnostic: does
performance on it tell us something about the model? The second is economic:
does buying more of it make the model better at what we care about? In practice
the two questions are often collapsed into a single loss-based notion of data
quality.

The answers can diverge. Consider a synthetic task whose loss falls steadily throughout
pretraining. This is compatible with at least three distinct situations. The
loss may be falling because general competence is rising and the task reads it
off; in that case the task is a useful instrument, while the value of added task
examples remains unknown. The loss may be falling because the model is being
shown the task and is fitting its surface form; in that case the task responds
to budget but the response may be entirely local. Or the loss may be falling
for both reasons while the source displaces something that the downstream target
actually needs; in that case buying more of it is actively harmful.

This paper separates the three signals and measures them in a single controlled
setting. We fix $70\%$ of the pretraining corpus to general Python source code
and vary only the composition of the remaining $30\%$, which we call the
\emph{controlled slice}. The slice is a mixture over three families: \oci{}
\citep{ahmad2025opencodeinstruct}, an instruction-formatted code corpus that is
close to the downstream target distribution; a \curated{} suite of $12$
synthetic tasks generated from a curated paper set to exercise code-adjacent
operations such as execution, search, state mutation, logical inference,
planning, ranking, and binding; and a \lit{} suite of $15$ synthetic probes
based on standard retrieval, state-tracking, composition, and transduction
tasks. Holding the base corpus, model family, training budget, and fine-tuning
stage fixed lets us attribute downstream differences to the slice alone.

Our contributions are as follows.

\begin{enumerate}
\item \textbf{Three separable signals.} We give operational definitions of
\emph{diagnosticity} (loss tracks global training progress),
\emph{teachability} (loss responds to the task's own token budget), and
\emph{transfer} (a source's share improves the downstream target), together
with an exposure-advantage diagnostic that compares task-specific exposure with
global pretraining step (\S\ref{sec:signals}).

\item \textbf{Teachability is common and family-dependent.} Across a per-task
budget sweep we detect teachability for $14$ of $27$ tasks. The split is
sharply asymmetric: $10$ of $12$ curated tasks are teachable, but only $4$ of
$15$ literature-derived probes are. False-discovery-rate adjustment retains
six curated and two literature-derived detections
(\S\ref{sec:task-results}).

\item \textbf{Transfer depends on mixture context.} On the
curated--\oci{} edge, \humaneval{} $\passk{20}$ is non-monotone in the curated
share, peaking at $22.6$ when \oci{} holds $75\%$ of the slice and falling to
$15.9$ at pure curated data---below the zero-synthetic baseline. The curated
suite is the most teachable family, while its pure-source endpoint falls below
the zero-synthetic baseline
(\S\ref{sec:mixture-results}).

\item \textbf{Loss-based scheduling follows the wrong signal.} \ado{}
\citep{jiang2024ado} allocates budget by predicted remaining loss reduction.
In every $60$k-step free-ratio run, \ado{} pushes the \oci{} share below $5\%$
within $5$k steps and to $1.2$--$1.4\%$ at the end of training. These runs trail
their mixture-matched fixed controls by $2.4$--$11.0$ percentage points of
$\passk{20}$ (\S\ref{sec:ado}).
\end{enumerate}

The practical rule we draw from this is narrow but concrete. When the downstream
target is code generation, keep the target-aligned source dominant in the
mixture, add curated synthetic data as a bounded share, and choose proportions
with a target-referenced signal.

\section{Related Work}

\paragraph{Data mixture optimization.}
DoReMi \citep{xie2023doremi} learns domain weights with a small proxy model
under a group-DRO objective. Mixture selection has also been posed as a
prediction problem \citep{ye2024datamixinglaws,liu2024regmix} and as online
reweighting driven by generalization estimates or skill structure
\citep{fan2024doge,chen2023skillit}. \ado{} \citep{jiang2024ado} fits
per-domain scaling laws during training and samples in proportion to predicted
remaining loss reduction, making it a direct instantiation of the reducibility
signal we isolate. Our contribution is a controlled demonstration that this
signal and downstream transfer can point in opposite directions.

\paragraph{Synthetic data for code.}
Instruction-formatted synthetic corpora are widely used in code-model training.
We use \oci{} \citep{ahmad2025opencodeinstruct} as aligned instruction data and
separate it from abstract synthetic task families. Our curated--\oci{} edge
makes the distinction visible: both families are generated by language models,
while \oci{} is closest to the downstream adaptation format.

\paragraph{Synthetic probes.}
A large literature builds controlled tasks to probe exact transduction, global
readout, retrieval, stack state, compositional interpretation, long-context
retrieval, and mutable state tracking. When these probes are also used as
training data, their diagnostic value leaves the effects of added exposure and
downstream transfer unestablished.

\paragraph{Code evaluation.}
We use \humaneval{} \citep{chen2021codex} as the downstream endpoint and report
the unbiased $\passk{k}$ estimator of \citet{chen2021codex}.

\section{Experimental Setup}
\label{sec:setup}

\subsection{Base corpus and controlled slice}

Every pretraining run draws $70\%$ of its tokens from a fixed general Python
corpus (\base{}) from CodeParrot and $30\%$ from a
\emph{controlled slice} whose composition is the only manipulated variable. A
run is therefore specified by a point $w = (w_O, w_C, w_L)$ on the $2$-simplex
$\Delta^2$, where the slice draws a fraction $w_O$ from \oci{}, $w_C$ from the
curated suite, and $w_L$ from the literature suite. The base fraction remains
fixed. The three slice weights sum to one, so increasing one family's share
strictly displaces the others:
\begin{equation}
P_r(x) = 0.70 P_{\mathrm{code}}(x) + 0.30\left[
w_{r,O}P_O(x)
+ w_{r,C}\sum_{t\in\mathcal T_C}w_{t\mid C}P_t(x)
+ w_{r,L}\sum_{t\in\mathcal T_L}w_{t\mid L}P_t(x)
\right].
\label{eq:mixture}
\end{equation}
Within-source task weights sum to one and remain fixed across runs. Each
machine-generated task cluster combines its difficulty variants with equal
problem weights before source-level aggregation.

\subsection{Source families}

\paragraph{OpenCodeInstruct.}
We use the pinned refined training split
\texttt{EER6/\allowbreak nvidia-\allowbreak OpenCodeInstruct-\allowbreak refined},
containing $444{,}611$ examples
\citep{ahmad2025opencodeinstruct}. This is the \emph{target-aligned} family:
its code-instruction response format matches the fine-tuning bridge and is
closest to the \humaneval{}-style downstream target. Controlled pretraining and
fine-tuning draw from this same refined split, so \oci{}-heavy pretraining is
additional exposure to the aligned instruction distribution.

\paragraph{Curated suite ($12$ tasks).}
Procedurally generated tasks inspired by a curated paper set. For each task, we
provide scraped paper text to a language model and request a self-contained
description of an automatically sampled task family. Each cluster has explicit
difficulty axes and deterministic answer generation; the full taxonomy is in
Appendix~\ref{app:tasks}.

\paragraph{Literature suite ($15$ tasks).}
Machine-generated versions of standard probes, including copy, reverse
sequence, majority, XOR parity, associative recall, Dyck completion, ListOps,
passkey retrieval, variable tracing, SCAN, three bAbI tasks, constant label,
and random label. They cover retrieval, state tracking, composition, and
transduction.

\subsection{Model and training}

All runs use a sparse-expert decoder-only model whose feed-forward blocks use
top-2 expert routing. The model has a $50{,}304$-token vocabulary, $12$ layers,
$12$ attention heads, width $768$, and SwiGLU width $2{,}048$. Training uses a
context length of $1{,}024$, effective batch size $256$, and $262{,}144$ tokens
per optimizer step. We train with AdamW at learning rate $10^{-3}$,
$\beta_1=0.9$, weight decay $10^{-4}$, and $500$ warmup steps. Runs last $60$k
or $120$k steps, with the decay horizon matched to run length. Full architecture
and scheduler settings appear in Appendix~\ref{app:hparams}.

The fixed-mixture shadow sweep preserves the $70/30$ split and varies the three
controlled sources along every pairwise edge in $25$-percentage-point steps.
The resulting $15$ mixtures define the downstream surface. Each synthetic task
appears in nine completed runs, with a median fourfold exposure range inside
the task-analysis window. One matched zero-synthetic baseline uses CodeParrot
alone.

\subsection{Downstream evaluation protocol}
\label{sec:eval}

Pretraining is followed by an identical fine-tuning stage for every run:
we fine-tune each available pretrained checkpoint for $1{,}000$ updates on the
refined \oci{} training split. Examples are shuffled and first-fit packed. We
use batch size $256$, Adam with $\beta_1=0.9$, $\beta_2=0.95$, and
$\epsilon=10^{-6}$, weight decay zero, $100$ linear warmup steps, cosine decay,
and peak learning rate $10^{-5}$. This common bridge keeps downstream
adaptation fixed across pretraining mixtures.

We evaluate the $164$-problem \humaneval{} test split
\citep{chen2021codex} with $20$ sampled completions per problem and report
$\passk{k}$ using the unbiased estimator
\begin{equation}
\widehat{\passk{k}} \;=\; \mathbb{E}_{\text{problems}}
\left[\, 1 - \frac{\binom{n-c}{k}}{\binom{n}{k}} \,\right],
\end{equation}
where $n=20$ and $c$ is the number of correct samples. The estimate is one when
$n-c<k$. For each value of $k$, the primary \emph{Best over steps} endpoint is
the highest score among fine-tuning checkpoints $100,200,\ldots,1000$. Every
paired bootstrap replicate resamples problems and repeats this checkpoint
selection. We report $\passk{5}$ and $\passk{10}$ alongside $\passk{20}$.

\section{Three Signals}
\label{sec:signals}

\subsection{Notation}

Let $\mathcal{T}$ be the set of $27$ synthetic tasks. For task $t$, run $r$,
and pretraining step $s$, let $L_{t,r}(s)$ be held-out validation loss. Let
$E_{t,r}(s)$ be cumulative loss-bearing answer tokens assigned to the task. The
task-analysis readout window is $s\in[5{,}000,55{,}000]$, and task exposure
$e_{t,r}$ is the change in $E_{t,r}$ between the first and last validation
readout in that window. Let $y(w)$ denote \humaneval{} $\passk{20}$ after
fine-tuning for a run with slice composition $w$.

Two properties of a task, and one property of a source, are distinguished by
which quantities they involve. Diagnosticity concerns global step $s$,
teachability concerns task-specific exposure $e_{t,r}$, and transfer concerns
the downstream endpoint $y$.

\subsection{Diagnosticity}

\begin{definition}[Diagnosticity]
Task $t$ is \emph{diagnostic} when its held-out loss tracks global pretraining
progress.
\end{definition}

We summarize this signal by regressing logged validation loss on global
pretraining step, pooling positive-loss readouts across the task's nine runs.
Its coefficient of determination is denoted $R^2_{t,\mathrm{step}}$. This
measurement asks how well a task reads out progress; the budget sweep separately
measures the role of examples from the task itself.

\subsection{Teachability}

\begin{definition}[Teachability]
Task $t$ is \emph{teachable} if its held-out loss responds to its own token
budget across runs.
\end{definition}

Teachability requires a cross-run comparison. We therefore run a \emph{task
budget sweep}: each task appears in nine fixed-mixture runs whose exposure
varies by a median factor of four. For task $t$ in run $r$, we first fit
\begin{equation}
\log L_{t,r}(s) = \alpha_{t,r} + \beta_{t,r}\log s + \epsilon_{t,r,s}
\end{equation}
over all readouts in the window and evaluate the fitted log-loss fall
$\Delta_{t,r}$ between its endpoints. We then fit
\begin{equation}
\Delta_{t,r} = a_t + b_t\log_2 e_{t,r} + \varepsilon_{t,r}.
\label{eq:teach}
\end{equation}
The slope $b_t$ reports learning per doubling of exposure, and $R^2_{t,\mathrm{budget}}$
is the share of across-run variation explained by budget. We shuffle exposures
across the nine runs $2{,}000$ times and declare a detection when the one-sided
permutation $p$-value is below $0.05$. We also report
Benjamini--Hochberg-adjusted counts.

\begin{remark}
Diagnosticity and teachability are logically independent. A task can combine
strong progress tracking with weak budget response, strong budget response with
weak progress tracking, strength on both signals, or weakness on both. To
compare the two explanations, we
define the exposure advantage
\begin{equation}
A_t = R^2_{t,\mathrm{exposure}} - R^2_{t,\mathrm{step}},
\end{equation}
where the first term regresses logged validation loss on cumulative exposure.
Seven tasks have $A_t>0$, and five of the six largest values are curated tasks.
\end{remark}

\subsection{Transfer, and why it is conditional}

\begin{definition}[Transfer]
A family $f$ \emph{transfers at $w$ relative to $w'$} when increasing its share
from the feasible mixture $w'$ to $w$ improves the downstream endpoint:
$y(w)>y(w')$.
\end{definition}

The definition is deliberately local because the simplex constrains every
increase to displace another source. We estimate transfer directly from
differences between completed mixtures on a pairwise edge and across the full
mixture surface. Transfer is a property of a source \emph{in a mixture
context}; this is what the title means by \emph{conditional}.

\subsection{The proxy-objective gap}

Adaptive schedulers optimize a surrogate. \ado{} \citep{jiang2024ado} fits
recent task-level loss curves and assigns greater sampling probability to tasks
with more predicted loss reduction remaining. In our configuration it updates
every $1{,}000$ steps using a $1{,}000$-step horizon. This is a teachability
signal measured online at the task level: it asks how much near-term loss
remains reducible. Its objective contains only pretraining loss.
Section~\ref{sec:ado} shows that the scheduler moves away from the region preferred by the downstream
surface for most of the $60$k-step training horizon.

\section{Task-Level Results}
\label{sec:task-results}

The fixed-mixture shadow runs make task budget the manipulated quantity. Runs
with nonzero curated or literature share contribute that source's tasks;
\oci{}-only and zero-synthetic runs omit synthetic tasks. Each
machine-generated task therefore appears in nine completed runs, and its
exposure inside the readout window varies by a median factor of four.

\begin{figure}[H]
\centering
\includegraphics[width=0.82\textwidth,height=0.42\textheight,keepaspectratio]
{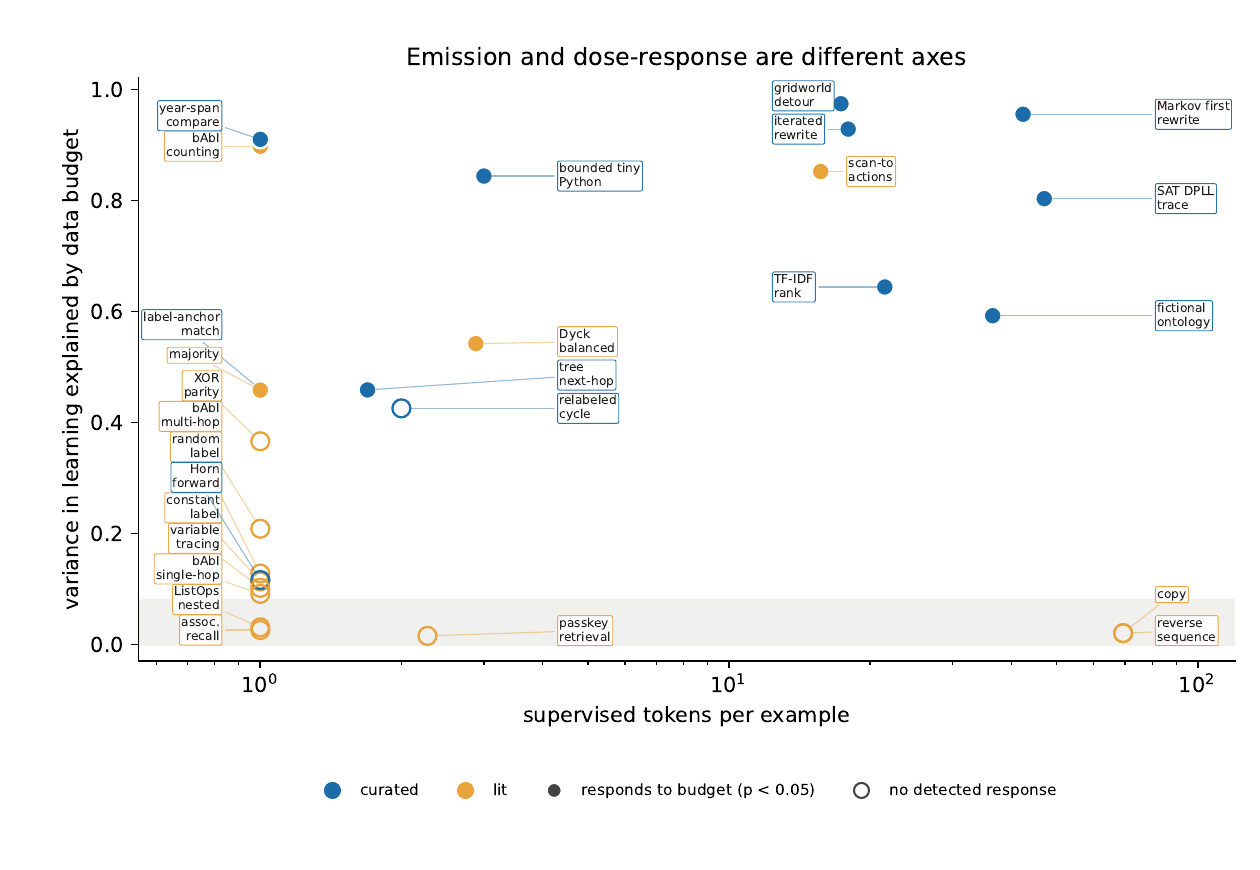}
\caption{Per-task response to scheduled token budget across fixed-mixture
shadow runs. The vertical axis is the share of across-run variation in fitted
log-loss fall explained by the task's own budget; the horizontal axis is
emitted tokens per example. Filled markers clear a permutation test against
shuffled budget, and the shaded band marks fits obtainable from shuffled
budget alone.}
\label{fig:task-budget}
\end{figure}

Fourteen of the $27$ tasks clear the permutation test: ten of twelve curated
tasks and four of fifteen literature-derived tasks. Benjamini--Hochberg
adjustment leaves eight detections, split six of twelve curated and two of
fifteen literature-derived. The family difference survives adjustment, while
the absolute detection count falls from fourteen to eight.

Curated tasks emit a median of $10.2$ supervised tokens per example, compared
with $1.0$ for literature probes, so emitted length could affect detection
power. Across all tasks, length is only weakly related to explained variance
(Pearson $r=0.34$, $p=0.08$; Spearman $\rho=0.27$, $p=0.18$). The extremes run
against a simple length explanation: copy and reverse sequence emit $69.3$
tokens per example and have $R^2=0.02$, while the one-token bAbI counting and
year-span comparison tasks respond strongly.

The exposure-advantage analysis provides a second view. Seven tasks have
$A_t>0$, meaning cumulative task exposure explains logged validation loss
better than global step. Five of the six largest advantages are curated tasks,
whereas seven of the ten lowest are literature-derived. A falling task loss can
therefore be a readout of general progress or a response to task-specific
exposure.

These associations come from a mixture sweep: a task's budget moves with the
surrounding source composition. Within this run set, curated code-adjacent
tasks more often link per-task budget to loss reduction, while several
literature-derived probes respond and some curated tasks have weak
task-specific evidence.

\section{Mixture-Level Transfer}
\label{sec:mixture-results}

\begin{table}[t]
\centering
\caption{\humaneval{} $\passk{20}$ after fine-tuning, along the
curated--\oci{} edge of the mixture simplex ($w_L = 0$). Values are Best over
steps percentages. The highest observed point has \oci{} dominant and a
minority curated share; pure curated data falls below the $16.5\%$
zero-synthetic baseline.}
\label{tab:edge}
\begin{tabular}{lccc}
\toprule
Slice composition & $w_O$ & $w_C$ & $\humaneval{}~\passk{20}$ \\
\midrule
Pure curated              & 0.00 & 1.00 & 15.9 \\
75\% curated              & 0.25 & 0.75 & 18.9 \\
50 / 50                   & 0.50 & 0.50 & 20.7 \\
25\% curated              & 0.75 & 0.25 & \textbf{22.6} \\
Pure \oci{}               & 1.00 & 0.00 & 19.5 \\
\bottomrule
\end{tabular}
\end{table}

\begin{figure}[H]
\centering
\includegraphics[width=\textwidth,height=0.46\textheight,keepaspectratio]
{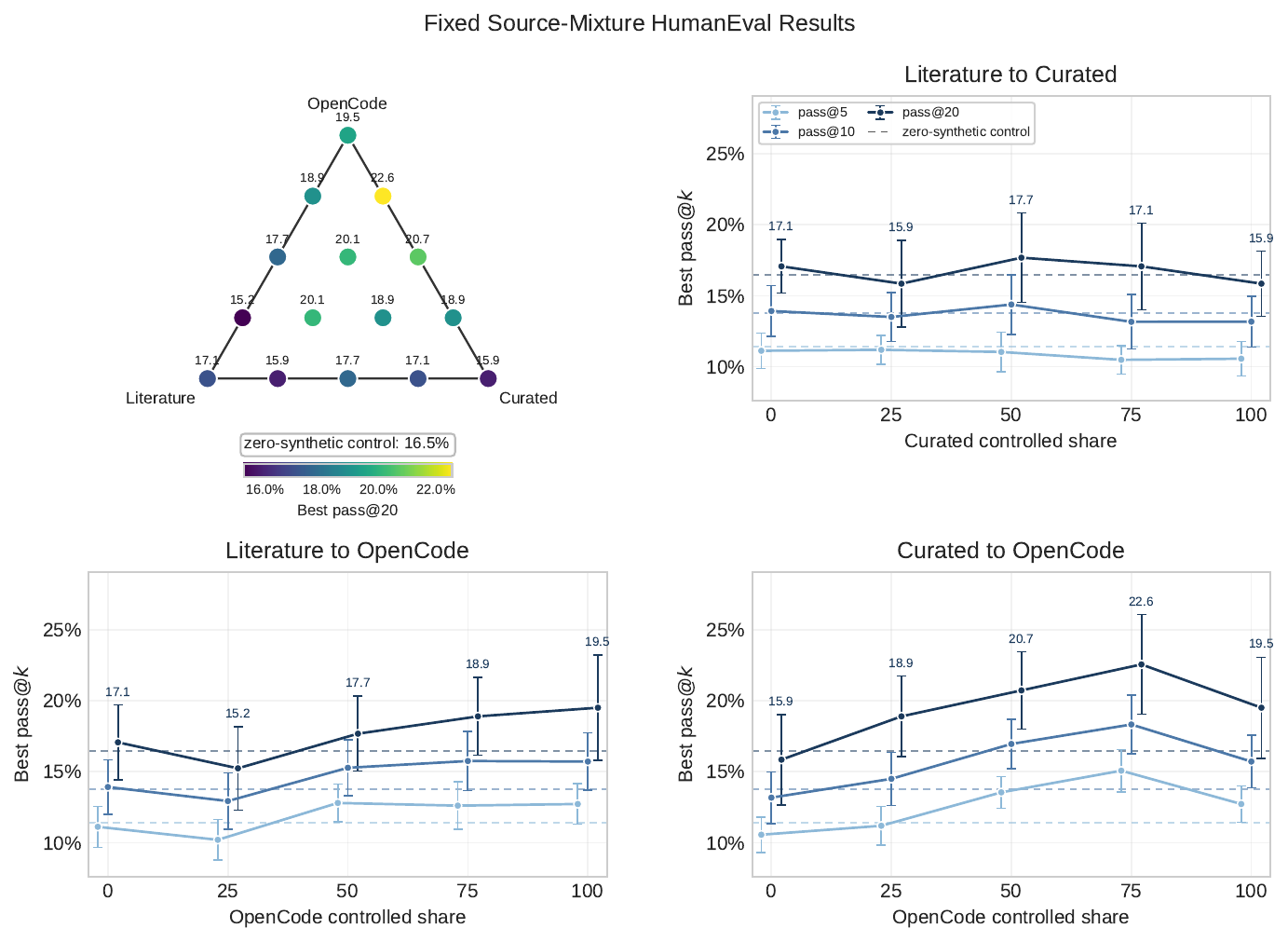}
\caption{Fixed-source-mixture \humaneval{} results for the shadow runs. The
ternary panel places each completed run by curated, literature, and \oci{}
shares, with color showing Best over steps $\passk{20}$. Edge panels show
$\passk{5}$, $\passk{10}$, and $\passk{20}$; annotated values are
$\passk{20}$. Dashed lines mark matched zero-synthetic baselines. Error bars
are $95\%$ within-problem intervals for comparing mixtures.}
\label{fig:mixture-space}
\end{figure}

\subsection{The curated--OpenCodeInstruct edge}

Table~\ref{tab:edge} contains the paper's central result. Moving along the edge
from pure curated data to pure \oci{}, $\passk{20}$ rises $15.9 \to 18.9 \to
20.7 \to 22.6$ and then falls to $19.5$. The highest observed point is interior
to the edge, at $75\%$ \oci{} / $25\%$ curated, and exceeds the pure-\oci{}
endpoint by $3.1$ percentage points.

The two ends of the edge tell opposite stories about the curated suite. At the
$25\%$ share it adds $3.1$ points relative to pure \oci{}. At the $100\%$ share
it scores $3.6$ points below pure \oci{} and $0.6$ points below the
zero-synthetic baseline. The ranking is consistent at $k=5,10,20$; at
$\passk{5}$, the best-observed mixture separates from both the neighboring
$50/50$ mixture and pure \oci{}.

Paired bootstrap intervals resample the $164$ problems $50{,}000$ times and
repeat Best-over-steps selection inside every replicate. They support an
\oci{}-dominant region with a limited curated share. A denser grid is required
to identify a precise optimum.

\subsection{Interpretation: capability versus alignment}

The pattern is consistent with complementarity between reusable program-like
operations and target-aligned instruction data. The curated tasks are teachable
(\S\ref{sec:task-results}) and exercise operations plausibly upstream of
program synthesis. \oci{} supplies code-instruction surface form close to the
fine-tuning bridge and downstream target. A bounded curated share can add those
operations while retaining an aligned source as most of the controlled slice.

This mechanism remains a hypothesis consistent with the data. Controlled
pretraining and fine-tuning use the same refined \oci{} split, so the
aligned-source effect includes repeated exposure to that instruction
distribution.

\subsection{Off-edge behavior and interaction structure}

Figure~\ref{fig:mixture-space} reports all $15$ completed fixed mixtures. The
ternary panel exposes the full surface, while the edge panels make the pairwise
tradeoffs visible at three values of $k$. The central claim is scoped to the
curated--\oci{} edge because its five reported points isolate the exchange
between the most teachable synthetic family and the aligned source.

\subsection{Teachability and transfer rank sources differently}

Putting \S\ref{sec:task-results} and this section together: the curated suite
has the highest teachability rate ($10/12$), yet its pure-source endpoint is
$15.9\%$, below both pure \oci{} and the zero-synthetic baseline. Teachability
therefore fails to order the observed downstream endpoints and is insufficient
as the sole mixture objective in this setup.

\section{Adaptive Scheduling Optimizes Reducibility and Loses Transfer}
\label{sec:ado}

The fixed sweep establishes that the transfer-favorable region is
\oci{}-dominant. We now ask what an online scheduler does when pointed at the
same three families.

Operationally, \ado{} makes the controlled-data weights in
Equation~\eqref{eq:mixture} time-dependent. For each controlled domain, it fits
a scaling law to recent loss measurements, extrapolates that curve over its
planning horizon, and estimates the loss reduction available from additional
tokens. Domains with the strongest predicted near-term loss reduction receive
greater sampling probability. Under preserved-ratio scheduling these proposals
change the within-source task weights; under free-ratio scheduling they also
change the source weights $(w_O,w_C,w_L)$ and therefore the location of the run
on the mixture simplex.

\subsection{Setup}

We run \ado{} \citep{jiang2024ado} over the controlled slice. The base $70\%$
remains fixed and unscheduled. \ado{} begins at step $1{,}000$, ignores the
first $500$ steps, and updates every $1{,}000$ steps over a $1{,}000$-step
horizon. It uses a minimum task probability of $0.01$, history coefficient
$0.9$, final smoothing $0.9$, and minimum scaling exponent $0.05$. We compare
two source-ratio policies:

\begin{itemize}
\item \textbf{Preserved ratio.} Source shares remain fixed while \ado{}
reallocates tasks within each source.
\item \textbf{Free ratio.} \ado{} reallocates controlled mass both within and
across sources.
\end{itemize}

Each $60$k-step, four-expert profile includes a fixed shadow run, a
preserved-ratio adaptive run, and a free-ratio adaptive run. Shadow mode records
the scheduler proposal while training on the fixed source mixture. A $120$k
free-ratio branch tests how the allocation changes beyond the $60$k comparison
horizon.

\subsection{The OpenCodeInstruct share collapses}

All three $60$k free-ratio runs drive the \oci{} share below $5\%$ of the
controlled slice within the first $5$k steps and finish at $1.2$--$1.4\%$.
They move mass before a downstream signal can appear and retain low \oci{}
shares through step $60$k. In the $120$k branches for the \oci{}-heavy and
curated-heavy profiles, the scheduler begins moving mass back around step
$90$k.

\begin{figure}[t]
\centering
\includegraphics[width=\textwidth]{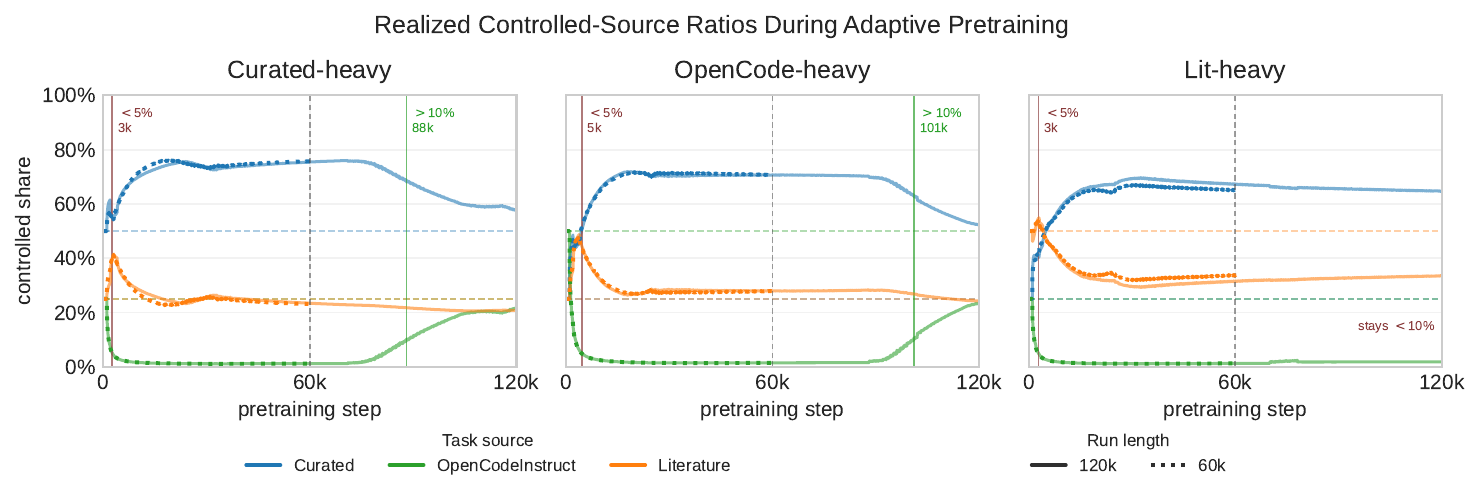}
\caption{Realized source ratios in adaptive free-ratio pretraining. Every run
abandons \oci{} within the first $5$k steps; recovery occurs only in the $120$k
branch, beyond the $60$k horizon used for downstream comparison. The learned
allocation is horizon-dependent.}
\label{fig:adaptive-source-ratio-drift}
\end{figure}

\subsection{Results}

\begin{table}[H]
\centering
\scriptsize
\caption{$60$k-step, four-expert mixture-profile and scheduler comparison.
Preserved ratio keeps source ratios at $50/25/25$, while free ratio reweights
sources. HumanEval columns report Best over steps $\passk{k}$. $\Delta$ shadow
compares $\passk{20}$ with the same-mixture shadow; final \oci{} share is
measured within the $30\%$ controlled slice.}
\label{tab:ado-results}
\begin{tabular}{lrrrrr}
\toprule
Run & $\passk{5}$ & $\passk{10}$ & $\passk{20}$ & $\Delta$ shadow & Final \oci{} share \\
\midrule
Zero-synthetic baseline  & 11.4\% & 13.8\% & 16.5\% & --- & --- \\
Curated-heavy shadow     & 12.7\% & 15.5\% & 18.9\% & +0.0 pp & 25.0\% \\
+ \ado{} preserved ratio & 11.0\% & 13.9\% & 17.1\% & -1.8 pp & 25.0\% \\
+ \ado{} free ratio      &  8.8\% & 12.4\% & 16.5\% & -2.4 pp & 1.2\% \\
\oci{}-heavy shadow      & 13.3\% & 16.5\% & 20.1\% & +0.0 pp & 50.0\% \\
+ \ado{} preserved ratio & 13.3\% & 16.0\% & 18.9\% & -1.2 pp & 50.0\% \\
+ \ado{} free ratio      &  3.7\% &  6.0\% &  9.1\% & -11.0 pp & 1.4\% \\
Literature-heavy shadow  & 13.6\% & 16.8\% & 20.1\% & +0.0 pp & 25.0\% \\
+ \ado{} preserved ratio & 11.0\% & 13.7\% & 17.1\% & -3.0 pp & 25.0\% \\
+ \ado{} free ratio      & 10.2\% & 12.7\% & 15.2\% & -4.9 pp & 1.2\% \\
\bottomrule
\end{tabular}
\end{table}

Preserved-ratio \ado{} keeps source budgets fixed and reaches
$17.1$--$18.9\%$ $\passk{20}$. Free-ratio \ado{} trails its matched shadow in
all three profiles, by $2.4$, $11.0$, and $4.9$ percentage points. Its
$\passk{20}$ endpoints range from $9.1\%$ to $16.5\%$, all below the
$22.6\%$ best-observed fixed mixture.

\subsection{Reading}

The mismatch follows from the scheduler's objective and time horizon. \ado{}
maximizes predicted near-term loss reduction over its tasks. \humaneval{} after
fine-tuning measures downstream transfer. In this setup, \oci{} scores low on
the first quantity early in training and high on the second, so the scheduler
moves mass away from the source preferred by the fixed-mixture surface.

The $1{,}000$-step fitting horizon also creates temporal myopia. The aligned
source abandoned in the first $5$k steps becomes attractive under the
scheduler's own criterion only after roughly $88$k steps. A transfer-aware
scheduler would incorporate a target-referenced marginal-value signal and
evaluate policies at several training horizons.

\section{Discussion}

\paragraph{Three questions, three measurements.}
Diagnosticity asks whether task loss tracks global progress and is measured by
the step-based fit. Teachability asks whether added task-specific exposure
predicts greater fitted loss reduction and is measured by the across-run budget
regression. Transfer asks whether changing a source share improves the
post-fine-tuning endpoint and is measured on the mixture surface. Each question
requires its own intervention or comparison.

\paragraph{Curated synthetic data has a bounded dose-response.}
The curated suite is genuinely useful: at a $25\%$ slice share it is worth $3.1$
$\passk{20}$ points over pure \oci{}. At a $100\%$ share it reaches $15.9\%$,
below the $16.5\%$ zero-synthetic baseline. Its value is bounded and
non-monotone, with the observed bound set by how much target-aligned data the
curated source displaces.

\paragraph{Standard probes are instruments.}
Four of fifteen literature-derived probes respond to their own budget, while
seven of the ten lowest exposure advantages also come from that family. These
tasks can remain useful controlled readouts of retrieval, state tracking,
composition, and transduction even when their own training budget has weak
evidence of marginal learning value.

\section{Limitations}
\label{sec:limitations}

\begin{itemize}
\item \textbf{Specific model family and base corpus.} The main comparisons use
a sparse-expert decoder trained with CodeParrot as the fixed $70\%$ source.
Other architectures, scales, and base corpora may produce a different mixture
surface.

\item \textbf{Joint format and content change.} \oci{} provides aligned
code-instruction responses, while the two task suites provide abstract
machine-generated problems. The experiment estimates source-family value;
format-specific and content-specific effects remain entangled.

\item \textbf{One downstream target.} The downstream evidence comes from
\humaneval{}. Transfer to other code benchmarks or general reasoning tasks
remains unmeasured.

\item \textbf{Coarse simplex resolution.} We sample the simplex at $15$ points.
The point $(0.75, 0.25, 0)$ is the best \emph{observed} point. Estimating an
argmax requires a denser grid.

\item \textbf{One scheduler.} We test \ado{} because its objective directly
instantiates reducibility. Other adaptive objectives and longer planning
horizons may yield different allocation trajectories.

\item \textbf{Mixture-coupled teachability estimate.} Each task has nine run
points, and its budget changes with the surrounding mixture. The permutation
test measures association across the completed grid; a task-isolated
intervention remains future work.

\item \textbf{Checkpoint selection.} Best over steps is optimistic by
$0.2$--$1.6$ percentage points across runs. The bootstrap repeats selection to
represent its uncertainty, while the endpoint remains an estimate of attainable
performance under checkpoint selection.
\end{itemize}

\section{Conclusion}

We separated three signals that are usually read off the same loss curve. Under
a controlled pretraining setup where only a $30\%$ slice varies, fourteen of
twenty-seven tasks respond detectably to their own exposure, and family-level
teachability ranks downstream code transfer differently. The best observed \humaneval{}
$\passk{20}$ came from a mixture in which the target-aligned source held $75\%$
of the slice and a highly teachable curated synthetic suite held the remaining
$25\%$; the same suite alone fell below the zero-synthetic baseline. A
loss-based adaptive scheduler, given the same three sources, drove the aligned
source below $5\%$ within the first $5$k steps and finished below all three
mixture-matched fixed controls. For code transfer,
keep the aligned source dominant, treat curated synthetic data as a bounded
share, and choose proportions with a target-referenced signal.

\bibliographystyle{plainnat}
\bibliography{refs}

\begin{thebibliography}{19}
\providecommand{\natexlab}[1]{#1}
\providecommand{\url}[1]{\texttt{#1}}
\expandafter\ifx\csname urlstyle\endcsname\relax
  \providecommand{\doi}[1]{doi: #1}\else
  \providecommand{\doi}{doi: \begingroup \urlstyle{rm}\Url}\fi

\bibitem[Ahmad et~al.(2025)Ahmad, Ficek, Samadi, Huang, Noroozi, Majumdar, and
  Ginsburg]{ahmad2025opencodeinstruct}
Wasi~Uddin Ahmad, Aleksander Ficek, Mehrzad Samadi, Jocelyn Huang, Vahid
  Noroozi, Somshubra Majumdar, and Boris Ginsburg.
\newblock Opencodeinstruct: A large-scale instruction tuning dataset for code
  llms.
\newblock \emph{arXiv preprint arXiv:2504.04030}, 2025.

\bibitem[Chen et~al.(2021)]{chen2021codex}
Mark Chen et~al.
\newblock Evaluating large language models trained on code.
\newblock \emph{arXiv preprint arXiv:2107.03374}, 2021.

\bibitem[Chen et~al.(2023)Chen, Roberts, Bhatia, Wang, Zhang, Sala, and
  R{\'e}]{chen2023skillit}
Mayee~F. Chen, Nicholas Roberts, Kush Bhatia, Jue Wang, Ce~Zhang, Frederic
  Sala, and Christopher R{\'e}.
\newblock Skill-it! a data-driven skills framework for understanding and
  training language models.
\newblock \emph{arXiv preprint arXiv:2307.14430}, 2023.

\bibitem[Dutta et~al.(2024)Dutta, Singh, Chakrabarti, and
  Chakraborty]{dutta2024thinkstep}
Subhabrata Dutta, Joykirat Singh, Soumen Chakrabarti, and Tanmoy Chakraborty.
\newblock How to think step-by-step: A mechanistic understanding of
  chain-of-thought reasoning, 2024.
\newblock URL \url{https://arxiv.org/abs/2402.18312}.

\bibitem[Fan et~al.(2024)Fan, Pagliardini, and Jaggi]{fan2024doge}
Simin Fan, Matteo Pagliardini, and Martin Jaggi.
\newblock Doge: Domain reweighting with generalization estimation.
\newblock \emph{arXiv preprint arXiv:2310.15393}, 2024.

\bibitem[Hanna et~al.(2023)Hanna, Liu, and Variengien]{hanna2023greaterthan}
Michael Hanna, Ollie Liu, and Alexandre Variengien.
\newblock How does gpt-2 compute greater-than?: Interpreting mathematical
  abilities in a pre-trained language model, 2023.
\newblock URL \url{https://arxiv.org/abs/2305.00586}.

\bibitem[Jiang et~al.(2024)Jiang, Zhou, Feng, Malladi, and
  Kolter]{jiang2024ado}
Yiding Jiang, Allan Zhou, Zhili Feng, Sadhika Malladi, and J.~Zico Kolter.
\newblock Adaptive data optimization: Dynamic sample selection with scaling
  laws.
\newblock \emph{arXiv preprint arXiv:2410.11820}, 2024.

\bibitem[Liu et~al.(2024)Liu, Zheng, Muennighoff, Zeng, Dou, Pang, Jiang, and
  Lin]{liu2024regmix}
Qian Liu, Xiaosen Zheng, Niklas Muennighoff, Guangtao Zeng, Longxu Dou, Tianyu
  Pang, Jing Jiang, and Min Lin.
\newblock Regmix: Data mixture as regression for language model pre-training.
\newblock \emph{arXiv preprint arXiv:2407.01492}, 2024.

\bibitem[Lyu et~al.(2026)Lyu, Huang, and Liang]{lyu2026surge}
Bohan Lyu, Siqiao Huang, and Zichen Liang.
\newblock Surge: On the potential of large language models as general-purpose
  surrogate code executors, 2026.
\newblock URL \url{https://arxiv.org/abs/2502.11167}.

\bibitem[Men et~al.(2024)Men, Cao, Jin, Chen, Liu, and Zhao]{men2024lookahead}
Tianyi Men, Pengfei Cao, Zhuoran Jin, Yubo Chen, Kang Liu, and Jun Zhao.
\newblock Unlocking the future: Exploring look-ahead planning mechanistic
  interpretability in large language models, 2024.
\newblock URL \url{https://arxiv.org/abs/2406.16033}.

\bibitem[Pan et~al.(2025)Pan, Ganesh, Abernethy, Esposo, and
  Lee]{pan2025satreasoning}
Leyan Pan, Vijay Ganesh, Jacob Abernethy, Chris Esposo, and Wenke Lee.
\newblock Can transformers reason logically? a study in sat solving, 2025.
\newblock URL \url{https://arxiv.org/abs/2410.07432}.

\bibitem[Salton and Buckley(1988)]{salton1988termweighting}
Gerard Salton and Christopher Buckley.
\newblock Term-weighting approaches in automatic text retrieval.
\newblock \emph{Information Processing \& Management}, 24\penalty0
  (5):\penalty0 513--523, 1988.

\bibitem[Saparov et~al.(2025)Saparov, Pawar, Pimpalgaonkar, Joshi, Pang,
  Padmakumar, Kazemi, Kim, and He]{saparov2025search}
Abulhair Saparov, Srushti Pawar, Shreyas Pimpalgaonkar, Nitish Joshi,
  Richard~Yuanzhe Pang, Vishakh Padmakumar, Seyed~Mehran Kazemi, Najoung Kim,
  and He~He.
\newblock Transformers struggle to learn to search, 2025.
\newblock URL \url{https://arxiv.org/abs/2412.04703}.

\bibitem[Todd et~al.(2026)Todd, Brinkmann, Gandikota, and
  Bau]{todd2026incontextalgebra}
Eric Todd, Jannik Brinkmann, Rohit Gandikota, and David Bau.
\newblock In-context algebra, 2026.
\newblock URL \url{https://arxiv.org/abs/2512.16902}.

\bibitem[Wang et~al.(2023)Wang, Li, Dai, Chen, Zhou, Meng, Zhou, and
  Sun]{wang2023labelanchors}
Lean Wang, Lei Li, Damai Dai, Deli Chen, Hao Zhou, Fandong Meng, Jie Zhou, and
  Xu~Sun.
\newblock Label words are anchors: An information flow perspective for
  understanding in-context learning, 2023.
\newblock URL \url{https://arxiv.org/abs/2305.14160}.

\bibitem[Xie et~al.(2023)Xie, Pham, Dong, Du, Liu, Lu, Liang, Le, Ma, and
  Yu]{xie2023doremi}
Sang~Michael Xie, Hieu Pham, Xuanyi Dong, Nan Du, Hanxiao Liu, Yifeng Lu, Percy
  Liang, Quoc~V. Le, Tengyu Ma, and Adams~Wei Yu.
\newblock Doremi: Optimizing data mixtures speeds up language model
  pretraining.
\newblock \emph{Advances in Neural Information Processing Systems}, 36, 2023.

\bibitem[Ye et~al.(2024)Ye, Liu, Sun, Zhou, Zhan, and
  Qiu]{ye2024datamixinglaws}
Jiasheng Ye, Peiju Liu, Tianxiang Sun, Yunhua Zhou, Jun Zhan, and Xipeng Qiu.
\newblock Data mixing laws: Optimizing data mixtures by predicting language
  modeling performance.
\newblock \emph{arXiv preprint arXiv:2403.16952}, 2024.

\bibitem[Zhang et~al.(2024)Zhang, Wang, and Charton]{zhang2024symboliccode}
Dylan Zhang, Justin Wang, and Francois Charton.
\newblock From symbolic tasks to code generation: Diversification yields better
  task performers, 2024.
\newblock URL \url{https://arxiv.org/abs/2405.19787}.

\bibitem[Zhao et~al.(2026)Zhao, Li, Bo, Takezoe, Hui, Guang, Ren, Qin, and
  Long]{zhao2026satquest}
Yanxiao Zhao, Yaqian Li, Zihao Bo, Rinyoichi Takezoe, Haojia Hui, Mo~Guang, Lei
  Ren, Xiaolin Qin, and Kaiwen Long.
\newblock Satquest: A verifier for logical reasoning evaluation and
  reinforcement fine-tuning of llms, 2026.
\newblock URL \url{https://arxiv.org/abs/2509.00930}.

\end{thebibliography}

\appendix

\section{Task Suite}
\label{app:tasks}

The controlled synthetic sources contain $15$ literature-derived task clusters
and $12$ curated task clusters. Each cluster combines its difficulty variants
with equal problem weights before source-level aggregation. \oci{} enters the
controlled mixture as aligned instruction data; the descriptions below cover
the machine-generated synthetic clusters.

\subsection{Literature-derived task clusters}

The literature-derived suite spans exact transduction, global readout,
retrieval, stack state, compositional interpretation, long-context retrieval,
and mutable state tracking. \textbf{Copy} and \textbf{reverse sequence} are exact
transduction tasks over digit sequences of length $16$, $64$, and $128$; copy
returns the input sequence unchanged, and reverse sequence returns it in reverse
order. \textbf{Majority} and \textbf{XOR parity} are global readout tasks over
binary sequences of length $16$, $64$, and $128$; majority outputs the bit that
occurs strictly more often in a tie-free sequence, while XOR parity outputs the
parity bit.

\textbf{Associative recall} tests retrieval and binding over $4$, $16$, or $64$
key-value pairs by asking for the value associated with a queried key.
\textbf{Dyck completion} tests stack state at depths $2$, $4$, and $8$ with two
bracket types by asking the model to complete a valid bracket prefix with the
closing brackets in stack order. \textbf{ListOps} tests compositional
interpretation at expression depths $2$, $4$, and $6$ by evaluating nested
prefix arithmetic expressions over small integer operators. \textbf{Passkey
retrieval} tests long-context retrieval at context lengths $128$, $512$, and
$1{,}024$ by asking for a five-digit passkey embedded in distractor context.

\textbf{Variable tracing} tests retrieval and binding with $1$, $2$, or $4$
assignment hops plus distractors by following assignment chains to return the
queried value. \textbf{SCAN command execution} tests compositional
interpretation on short, medium, and long commands by translating navigation
commands into primitive action tokens. The three bAbI clusters test mutable
state tracking with $5$ or $20$ story facts: \textbf{bAbI single-hop} tracks the
latest location of a named person, \textbf{bAbI multi-hop} combines an ownership
fact with a movement fact to infer an object's location, and \textbf{bAbI
counting} accumulates object-count changes to return the queried total.
\textbf{Constant label} emits a fixed answer for every input, and \textbf{random
label} emits a label sampled independently of the input from a two-label set.

\subsection{Curated task clusters}

The curated suite emphasizes program-like operations, formal inference,
planning, ranking, and in-context binding. \textbf{Bounded tiny Python
execution} is a surrogate code-execution task whose difficulty varies by
variables, updates, loop depth, loop iterations, and branch rate; the model
executes a restricted Python program and predicts the printed values. The task
follows the SURGE setting by asking for outputs of small bounded programs from
the source text alone \citep{lyu2026surge}.

\textbf{Fictional ontology reasoning} is a relational-reasoning task whose
difficulty varies by inference hops and distractor facts; the model answers a
property query from shuffled ontology facts with a canonical reasoning chain.
It mirrors the PrOntoQA-style fictional ontology setting, where solving requires
selecting relevant facts and composing the support chain
\citep{dutta2024thinkstep}. \textbf{Gridworld detour planning} is a
lookahead-planning task whose difficulty varies by grid size, path length,
branch count, and branch length; the model finds the unique shortest action
sequence through a gridded maze with detours. It captures the look-ahead
planning setup in a compact fully observed domain where the shortest plan
depends on anticipating later states \citep{men2024lookahead}.

\textbf{Horn forward chaining} is a logical-forward-chaining task whose
difficulty varies by derivation hops and distractor rules; the model applies
Horn-style implication rules to decide whether a queried atom follows from the
facts. It follows SATQuest's verifier-centered logical-reasoning setup with
objective labels, automatic checking, proof depth, and irrelevant rule
distractors \citep{zhao2026satquest}. \textbf{Iterated string replace} is a
mutable-state-tracking task whose difficulty varies by initial length,
replacement steps, alphabet size, and rewrite constraints; the model runs an
ordered program of string replacements and returns the final string. It draws
from the symbolic string-replacement setting used for Markov-algorithm tasks,
using ordered Python global replacements as the sampled program
\citep{zhang2024symboliccode}. \textbf{Label-anchor feature match} is an
in-context-binding task whose difficulty varies by class count, shots per
class, and item length; the model uses in-context examples to map a feature
marker to an arbitrary label. It makes demonstration label words the arbitrary
anchors that bind sampled feature markers to final predictions, matching the
label-token information-flow account of in-context learning
\citep{wang2023labelanchors}.

\textbf{Markov first rewrite} is an algorithmic string-rewriting task whose
difficulty varies by rule count, input length, rule length, and no-op rate; the
model applies the first matching rewrite rule once or leaves the string
unchanged. It follows the Markov-algorithm string-replacement setup with fresh
ordered rule lists, instruction selection, and exact substring replacement
\citep{zhang2024symboliccode}. \textbf{Relabeled cyclic row completion} is an
in-context-algebra task whose difficulty varies by cyclic order and distractor
facts; the model completes a missing product in a relabeled cyclic operation
table. It isolates the in-context variable-token algebra setup, where visible
symbols receive fresh per-example meanings and the answer follows from the
local cyclic-group row \citep{todd2026incontextalgebra}. \textbf{SAT DPLL trace}
is a formal-logical-reasoning task whose difficulty varies by variable count,
clause count, and maximum trace length; the model produces a deterministic DPLL
trace for a small 3-CNF formula. It follows the SAT-solving-with-chain-of-thought
setup by requiring the deductive DPLL reasoning path over formula sizes small
enough for exact generation \citep{pan2025satreasoning}.

\textbf{TF-IDF document ranking} is a weighted-retrieval-ranking task whose
difficulty varies by document count, document length, vocabulary size, and query
terms; the model ranks synthetic documents under an explicit term-frequency and
rarity scoring rule. It operationalizes classical term-frequency and inverse
document-frequency weighting as a replayable synthetic ranking problem with
deterministic tie-breaking \citep{salton1988termweighting}. \textbf{Tree
next-hop search} is a search-and-planning task whose difficulty varies by
branching factor and depth; the model returns the next edge on the path from a
start node to a goal node in a serialized tree. It is a controlled next-hop
search problem that requires identifying the first step on a path from the
visible graph, start, and reachable goal \citep{saparov2025search}.
\textbf{Year-span comparison} is a numerical-comparison task whose difficulty
varies by minimum and maximum year gap; the model decides whether the end year
in a sentence is later than the start year. It turns the year-span greater-than
continuation setting into a deterministic binary classification probe with
controlled suffix gaps and one unambiguous answer token
\citep{hanna2023greaterthan}.

\section{Task-Level Analysis Details}
\label{app:task-analysis}

\paragraph{Readout window.}
Both task-level diagnostics use pretraining steps $5{,}000$ through $55{,}000$.

\paragraph{Task exposure.}
Cumulative loss-bearing answer tokens received by a task, differenced between
the first and last validation readouts in the window. The training objective is
applied only to answer tokens, making this the exposure on which the task
trains.

\paragraph{Fitted log-loss fall.}
For one task in one run, we fit $\log L$ against $\log s$ over every validation
readout in the window and evaluate the fitted difference between the window
endpoints. The fit limits the influence of any single noisy readout.

\paragraph{Per-task budget statistic.}
For each task, we regress fitted log-loss fall on $\log_2$ exposure across its
nine runs. The slope is learning per doubling of exposure, and $R^2$ is the
share of across-run variation explained by exposure.

\paragraph{Permutation calibration.}
We shuffle exposures across the nine runs within a task, refit the regression,
and take the one-sided $p$-value as the fraction of $2{,}000$ permutations
reaching at least the observed $R^2$. Detection uses unadjusted $p<0.05$;
Benjamini--Hochberg-adjusted counts are reported alongside.

\paragraph{Exposure advantage.}
For each task, exposure advantage is the $R^2$ of logged validation loss
regressed on cumulative task exposure minus the $R^2$ from regression on global
pretraining step. Both regressions pool every positive-loss readout across the
task's nine runs.

\paragraph{Emitted-length check.}
Curated tasks emit a median of $10.2$ supervised tokens per example, compared
with $1.0$ for literature probes. Across all $27$ tasks, emitted length and
explained variance are weakly related (Pearson $r=0.34$, $p=0.08$; Spearman
$\rho=0.27$, $p=0.18$). Copy and reverse sequence emit $69.3$ tokens per
example and have $R^2=0.02$, while bAbI counting and year-span comparison emit
one token and respond strongly.

\section{Training Hyperparameters}
\label{app:hparams}

Experimental runs use the $70/30$ split between CodeParrot and controlled data,
while matched zero-synthetic baselines use only CodeParrot. CodeParrot uses
\texttt{codeparrot\_train\_v2\_near\_dedup\_safe}; \oci{} uses
\texttt{EER6/nvidia-OpenCodeInstruct-refined}. The pinned refined train split
contains $444{,}611$ examples. Controlled-data profiles emphasize curated,
\oci{}, or literature data with weights $50/25/25$. Synthetic examples are
grouped by task cluster and sampled with mixture temperature $1.0$.

Models use a $50{,}304$-token vocabulary, $12$ layers, $12$ heads, width $768$,
and SwiGLU width $2{,}048$. Sparse-expert feed-forward blocks use top-2 routing;
the expert cap is disabled. Each token routes to two experts, so expert count changes
conditional capacity and routing competition while active route count stays
fixed. Training uses context length $1{,}024$, effective batch size $256$, and
$262{,}144$ tokens per optimizer step. Model parameters are replicated. AdamW
uses learning rate $10^{-3}$, $\beta_1=0.9$, weight decay $10^{-4}$, and $500$
warmup steps. Runs last $60$k or $120$k steps, with decay matched to run length.

All runs process $262{,}144$ tokens per update, totaling $15.73$B tokens at
$60$k steps and $31.46$B at $120$k steps. Across four-, eight-, and
sixteen-expert models, the shared backbone remains fixed at $66.947$M
parameters and top-2 routing activates $113.246$M expert parameters per token.
Stored parameters increase from $293.476$M to $520.005$M and $973.064$M, while
active parameters remain approximately $180$M per token.

\section{ADO Configuration}
\label{app:ado}

\ado{} is applied only to the $30\%$ controlled slice; the $70\%$ base
fraction is fixed. It begins at step $1{,}000$, ignores the first $500$ steps,
and updates every $1{,}000$ steps over a $1{,}000$-step horizon. It uses a
minimum task probability of $0.01$, history coefficient $0.9$, final smoothing
$0.9$, and minimum scaling exponent $0.05$. The objective uses answer tokens.
Preserved-ratio adaptive runs reallocate within each source, and free-ratio runs
also reallocate mass across sources.

\section{Evaluation Details}
\label{app:eval-details}

\paragraph{Fine-tuning.}
Each pretrained checkpoint is fine-tuned on the default subset of the refined
\oci{} training split. Examples are shuffled and packed using a first-fit
strategy. We use batch size $256$ and Adam with $\beta_1=0.9$, $\beta_2=0.95$,
and $\epsilon=10^{-6}$. Training lasts $1{,}000$ updates with $100$ linear
warmup steps followed by cosine decay and peak learning rate $10^{-5}$; weight
decay is zero.

\paragraph{HumanEval evaluation.}
The OpenAI \humaneval{} test split contains $164$ problems. For each problem,
we sample $20$ completions using a $2{,}048$-token evaluation context and at
most $512$ generated tokens. Decoding uses temperature $1.0$,
\texttt{top\_k=-1}, and \texttt{top\_p=1.0}. Generation stops at an EOS token
or the generation limit. Each completion is executed in an isolated Python
environment and is correct when it passes the corresponding tests.

\paragraph{Checkpoint aggregation.}
We evaluate fine-tuning checkpoints from steps $100$ through $1{,}000$. The
primary endpoint, \emph{Best over steps}, is the highest observed $\passk{k}$
across these checkpoints. Selection inflates the endpoint by $0.2$--$1.6$
percentage points depending on the run, and a bootstrap replicate re-selects
the observed best checkpoint $32\%$--$73\%$ of the time.

\paragraph{Bootstrap and interval conventions.}
For the primary mixture edge, paired bootstrap intervals use $50{,}000$
replicates. Each replicate resamples the $164$ problems and repeats
Best-over-steps checkpoint selection. Problems are drawn once per replicate and
shared across runs, keeping the comparison paired.

Figure intervals for mixture comparisons use $95\%$ within-problem intervals
with the Morey bias correction. Every mixture is scored on the same $164$
problems, so common problem difficulty shifts the panel together. Removing that
common component yields intervals suited to comparisons among mixtures.

\paragraph{Reporting conventions.}
We report $\passk{k}$ as a percentage rounded to one decimal place and
differences in percentage points.

\begin{table}[H]
\centering
\scriptsize
\caption{\humaneval{} Best over steps uncertainty for the primary four-expert
shadow-source comparison. Counts are problem-level $\passk{20}$ solved
indicators over $164$ problems. W/L/T gives paired wins, losses, and ties versus
the matched zero-synthetic baseline. Intervals are $95\%$ paired bootstrap
intervals from $50{,}000$ deterministic resamples.}
\label{tab:humaneval-uncertainty}
\begin{tabular}{lrrrrr}
\toprule
Run & Solved & W/L/T & $\Delta\passk{20}$ & $95\%$ CI & $P(\Delta>0)$ \\
\midrule
Curated-heavy shadow & 31/164 vs 27/164 & 9/5/150 & +2.4 pp & [-1.8, +6.7] pp & 82.7\% \\
\oci{}-heavy shadow & 33/164 vs 27/164 & 10/4/150 & +3.7 pp & [-0.6, +7.9] pp & 93.3\% \\
Literature-heavy shadow & 33/164 vs 27/164 & 11/5/148 & +3.7 pp & [-1.2, +8.5] pp & 91.9\% \\
\bottomrule
\end{tabular}
\end{table}

\section{Curated Task Construction Prompt}
\label{app:curated-task-prompt}

The curated task clusters were elicited one at a time with the following
self-contained prompt:
\begingroup
\footnotesize
\begin{verbatim}
I need a synthetic task that can be automatically sampled (in the format shown after the paper).
please suggest one such task.
answer in prose, in a self contained fashion, in a way someone could implement
the task based on your description.

{paper_text}

<example>

_ALPHABET = ("0", "1")

@register_task
class XorParityTask(SyntheticTask):
    task_id = "xor_parity"

    def sample(self, rng: random.Random, split: str, difficulty: dict[str, Any]) -> Example:
        tokens = [rng.choice(_ALPHABET) for _ in range(length)]
        parity = 0
        for tok in tokens:
            parity ^= int(tok)
        answer = str(parity)
        sequence = " ".join(tokens)
        prompt = f"<task=xor_parity>\nInput: {sequence}\nAnswer:"
        target = f" {answer}"
        full_text = prompt + target
        loss_mask = build_loss_mask(prompt, target)
        return Example(
            task_id=self.task_id,
            domain_id=f"xor_parity_len_{length}",
            prompt=prompt,
            target=target,
            full_text=full_text,
            loss_mask=loss_mask,
            metadata={
                "task_id": self.task_id,
                "macro_family": "finite_state_or_global_readout",
                "difficulty": dict(difficulty),
            },
        )

</example>
\end{verbatim}
\endgroup

\section{Provenance of Reported Numbers}
\label{app:provenance}

The main empirical quantities map to the paper as follows. The five
curated--\oci{} $\passk{20}$ values appear in Table~\ref{tab:edge}, and the
full $15$-mixture surface and lower-$k$ endpoints appear in
Figure~\ref{fig:mixture-space}. Task-budget detections use the definitions in
Appendix~\ref{app:task-analysis} and are visualized in
Figure~\ref{fig:task-budget}. The $60$k scheduler endpoints and final source
shares appear in Table~\ref{tab:ado-results}; their trajectories appear in
Figure~\ref{fig:adaptive-source-ratio-drift}. Evaluation selection and interval
conventions are specified in Appendix~\ref{app:eval-details}.

\end{document}